\documentclass[10pt,twocolumn,letterpaper]{article}

\usepackage[pagenumbers]{cvpr} % To force page numbers, e.g. for an arXiv version

\usepackage[dvipsnames]{xcolor}

\definecolor{cvprblue}{rgb}{0.21,0.49,0.74}
\usepackage[pagebackref,breaklinks,colorlinks,citecolor=cvprblue]{hyperref}

\usepackage{hyperref}
\usepackage{url}
\usepackage{graphicx}
\usepackage{wrapfig}
\usepackage{algorithm}
\usepackage{algorithmic}
\usepackage{bm}
\usepackage{multicol}
\usepackage{multirow}
\usepackage[accsupp]{axessibility}

\newtheorem{theorem}{Theorem}
\newcommand{\hong}[1]{{\color{black}#1}\normalfont}

\def\paperID{7130} % *** Enter the Paper ID here
\def\confName{CVPR}
\def\confYear{2024}

\title{FedMef: Towards Memory-efficient Federated Dynamic Pruning}

\author{Hong Huang$^{1}$ \enspace Weiming Zhuang$^{2}$ \enspace Chen Chen$^{2}$ \enspace Lingjuan Lyu$^{2}$\thanks{corresponding author} \\
$^{1}$City University of Hong Kong \enspace $^{2}$ Sony AI \\
{\tt\small hohuang@cityu.edu.hk} \enspace {\tt\small \{weiming.zhuang, chena.chen, lingjuan.lv\}@sony.com}
}

\begin{document}
\maketitle
\begin{abstract}
Federated learning (FL) promotes decentralized training while prioritizing data confidentiality. However, its application on resource-constrained devices is challenging due to the high demand for computation and memory resources to train deep learning models. Neural network pruning techniques, such as dynamic pruning, could enhance model efficiency, but directly adopting them in FL still poses substantial challenges, including post-pruning performance degradation, high activation memory usage, etc. To address these challenges, we propose FedMef, a novel and memory-efficient federated dynamic pruning framework. FedMef comprises two key components. First, we introduce the budget-aware extrusion that maintains pruning efficiency while preserving post-pruning performance by salvaging crucial information from parameters marked for pruning within a given budget. Second, we propose scaled activation pruning to effectively reduce activation memory footprints, which is particularly beneficial for deploying FL to memory-limited devices. Extensive experiments
demonstrate the effectiveness of our proposed FedMef. In particular, it achieves a significant reduction of 28.5\% in memory footprint compared to state-of-the-art methods while obtaining superior accuracy.
\end{abstract}
\section{Introduction}
% \vspace{-0.2cm}
Federated learning (FL) has emerged as an important paradigm for the training of machine learning models %on decentralized devices
across decentralized clients while preserving the confidentiality of local data~\cite{mcmahan2017communication, li2019convergence, zhuang2023mas}. In particular, cross-device FL, as described in~\cite{kairouz2021advances}, places emphasis on scenarios where FL clients predominantly consist of edge devices with resource constraints. Cross-device FL has gained significant attention in academic research and industry applications, fueling a wide range of applications, including Google Keyboard~\cite{hard2018federated,leroy2019federated}, Apple Speech Recognition~\cite{paulik2021federated}, etc.
Despite its success, the resource-intensive nature of training models, which includes high computational and memory costs, poses challenges for the deployment of cross-device FL on resource-constrained devices.

Neural network pruning~\cite{janowsky1989pruning, han2015deep, molchanov2019importance, singh2020woodfisher} is a potential solution to improve the efficiency of the model and reduce the high demand for resources. 
However, a closer inspection of some previous work on the application of neural network pruning to FL~\cite{shao2019privacy, li2021lotteryfl, liu2021adaptive, munir2021fedprune} reveals a potential pitfall: They often rely on initial training of dense models, similar to centralized pruning methodologies~\cite{han2015deep, molchanov2019importance, singh2020woodfisher}.
% \wm{revise the last sentence: hard to understand what is the shortcoming - high memory? high computation? low accuracy?.} 
These federated pruning methods are not suitable for cross-device FL because the training of dense models still requires high computation and memory costs on resource-constrained devices.
\begin{figure*}[t]
  \centering
  \includegraphics[width=0.95\linewidth]{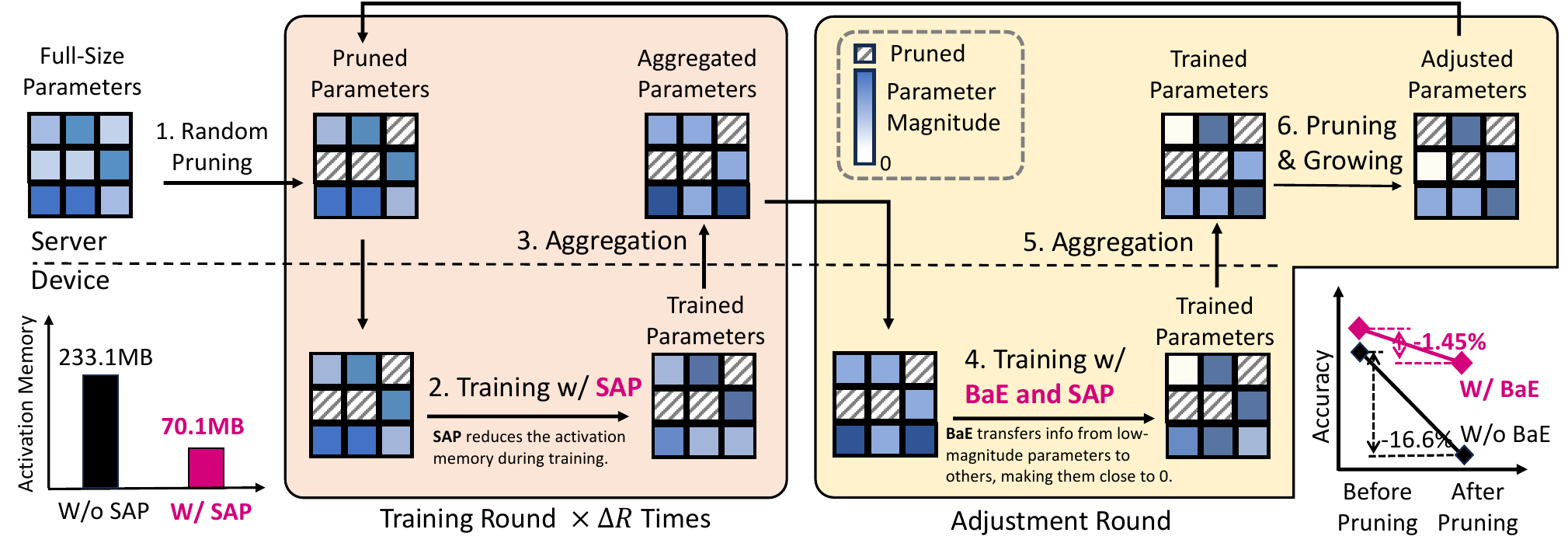}
  \vspace{-0.3cm}
  \caption{ Overview of FedMef for the memory-efficient dynamic pruning in federated learning. FedMef proposes budget-aware extrusion (BaE) to preserve post-pruning accuracy by transferring essential information from low-magnitude parameters to the others, making them close to 0, and introducing scaled activation pruning (SAP) to reduce memory usage. In FedMef, the server distributes a randomly pruned model to devices for collaborative training with SAP. After multiple training rounds, devices employ BaE for information transfer. The server adjusts the model structure through magnitude pruning and growing. The newly activated parameters are initialized as 0.
    }
  \label{fig:framework}
  \vspace{-0.3cm}
\end{figure*}

\begin{figure}[t]
  \centering
  \includegraphics[width=\linewidth]{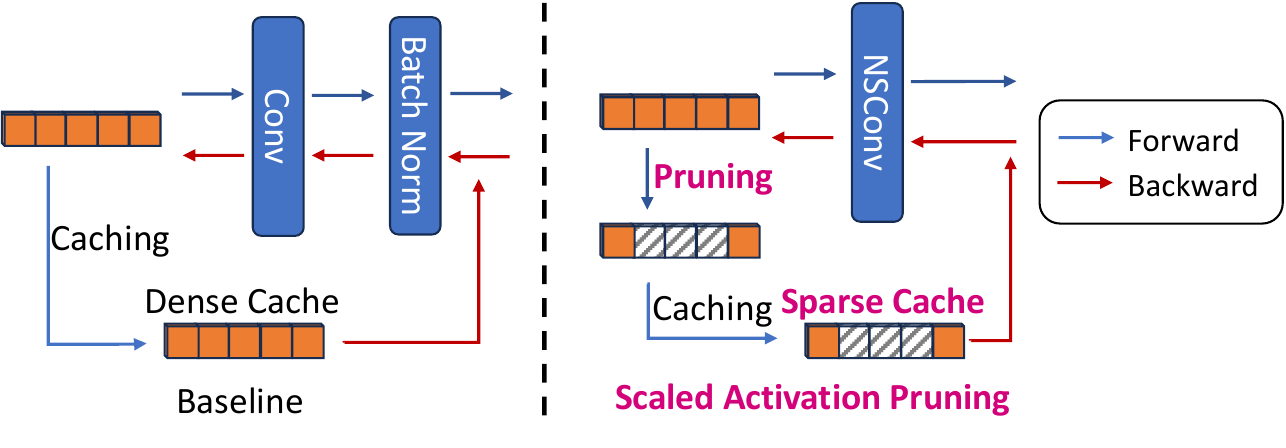}
  % \vspace{-0.3cm}
  \caption{ \hong{The illustration of training pipeline in baseline and the proposed scaled activation pruning method. During the forward pass, the scaled activation pruning generates near-zero activation via the Normalized Sparse Convolution (NSConv). Then, the dense activation caches are pruned based on magnitude. During the backward pass, these pruned caches are used to compute the gradients. Scaled activation pruning significantly saves activation memory footprints by more than 3 times in the CIFAR-10 dataset with the MobileNetV2 model.
   }}
  \vspace{-0.3cm}
  \label{fig:act_prune}
\end{figure}

To address these challenges, recent research has shifted to federated dynamic pruning~\cite{jiang2022model, qiu2022zerofl, bibikar2022federated, huang2022fedtiny}. 
These frameworks derive specialized pruned models by iterative adjustment of sparse on-device models. Devices start with a randomly pruned model, followed by traditional FL training, and periodically adjust the sparse model structure through pruning and growing operations~\cite{evci2020rigging}. 
Through iterative training and adjustments, devices can develop specialized pruned models bypassing the need to train dense models, which reduces both computational and memory demands.

However, existing federated dynamic pruning frameworks~\cite{jiang2022model, qiu2022zerofl, bibikar2022federated, huang2022fedtiny} face two issues: significant post-pruning accuracy degradation and substantial activation memory usage. First, these frameworks cause a significant decline in accuracy after magnitude pruning because they hastily eliminate low-magnitude parameters, regardless of the substantial information they may contain. Such incautious parameter pruning often results in the model's inability to regain its previous accuracy before the subsequent pruning iteration, ultimately leading to suboptimal end-of-training performance. Second, these frameworks fail to reduce the memory footprint of activation. For certain widely adopted models for edge deployment, like MobileNet~\cite{sandler2018mobilenetv2}, %frequently employed for edge deployment, 
a significant portion of the total memory footprints is allocated to activation. However, current federated dynamic pruning methods focus primarily on reducing the model size, overlooking optimization for activation memory.
% In certain models (\textit{e.g.,} MobileNet~\cite{sandler2018mobilenetv2}), the memory reserved for activation constitutes a significant fraction of the overall memory allocation. Unfortunately, the prevailing approaches fail to provide memory optimizations in this crucial segment.

In this work, we introduce FedMef, an \textbf{Fed}erated \textbf{M}emory-\textbf{ef}fcient dynamic pruning framework that adeptly addresses all the aforementioned challenges. Figure~\ref{fig:framework} illustrates the workflow of FedMef and highlights our two new proposed components. First, FedMef presents \textbf{\textit{budget-aware extrusion} (BaE)} to address the challenge of post-pruning accuracy degradation.
% , as depicted in Figure~\ref{fig:framework}. To address the challenge of post-pruning accuracy degradation, FedMef proposes \textit{budget-aware extrusion}. 
Rather than simply discarding low-magnitude parameters, our method salvages essential information from these potential pruning candidates by transferring it to other parameters through a surrogate loss function within a preset budget. 
Second, FedMef proposes \textbf{\textit{scaled activation pruning} (SAP)} to address the problem of high activation memory.
% To tackle the activation memory problem, FedMef incorporates the \textit{scaled activation pruning} technique, as illustrated in Figure~\ref{fig:act_prune}. 
This method performs activation pruning during the training process to dramatically reduce the memory footprints of the activation caches, as illustrated in Figure~\ref{fig:act_prune}. To enhance the efficacy of SAP, especially for devices with severe memory constraints, we are inspired by recent methods that remove batch normalization (BN) layers~\cite{brock2021characterizing, zhuang2023normalization} and 
% remove conventional BN layers and 
replace convolution layers with Normalized Sparse Convolution (NSConv). NSConv can normalize most of the activation elements to be or close to zero. This reduces the disparity between original and pruned activation, mitigating the degradation of accuracy in SAP. 
% To optimize the efficacy of scaled activation pruning, particularly for severely memory-restricted devices operating with minuscule batch sizes, inspried by recent batch normalization free methods~\cite{brock2021characterizing, zhuang2023normalization}, we %jettison
% abandon the traditional batch normalization layers in favor of the Weight Sparse Standardization convolution (WSSConv). WSSConv's unique capability to normalize the majority of activation map elements to zero or near zero, independent of batch size, curtails accuracy losses during the scaled activation pruning.

We conducted extensive experiments on three datasets: CIFAR-10~\cite{krizhevsky2009learning}, CINIC-10~\cite{darlow2018cinic}, and TinyImageNet~\cite{le2015tiny}, using ResNet18~\cite{he2016deep} and MobileNetV2~\cite{sandler2018mobilenetv2} models. Extensive experimental results suggest that FedMef outperforms the state-of-the-art (SOTA) methods on all datasets and models. In addition, FedMef requires fewer memory footprints than SOTA methods. For example, FedMef significantly reduces the memory footprint of MobileNetV2 by 28.5\% while improving the accuracy by more than 2\%.

\section{Related Work}
\subsection{Neural Network Pruning}
Neural network pruning, which emerged in the late 1980s, aims to reduce redundant parameters in deep neural networks (DNNs). Traditional techniques focus on achieving a trade-off between accuracy and sparsity during inference. This typically involves assessing the importance of parameters in a pre-trained DNN and discarding those with lower scores. Various methods are employed to determine these scores, such as weight magnitude~\cite{janowsky1989pruning, han2015deep} and Taylor expansion of the loss functions~\cite{mozer1988skeletonization, lecun1989optimal, molchanov2019pruning}. 
However, these methods need
% A major drawback of these methods is the necessity 
to train a dense model first, which increases both computational and memory costs.

Modern pruning has shifted its focus towards enhancing the efficiency of DNN training processes. %This newer approach, known as
For example, dynamic sparse training~\cite{mocanu2018scalable,dettmers2019sparse, evci2020rigging}, actively adjusts the architecture of the pruned model throughout the training while maintaining desired sparsity levels. Nevertheless, these methods only simply prune low-magnitude parameters %and %do not address
while neglecting the memory consumption of the activation caches, resulting in decreased accuracy and sub-optimal memory optimization. 
% Such oversights can culminate in diminished accuracy and suboptimal memory optimization.

% Since the late 1980s, pruning neural networks has been a popular way to reduce the number of unnecessary parameters in a deep neural network (DNN). Most existing pruning techniques focus on the balance between accuracy and sparsity during the inference stage. This process usually starts by calculating the importance scores of all parameters in a pretrained DNN, then eliminating those with lower scores. The importance scores can be determined based on the magnitude of the weights~\cite{janowsky1989pruning, han2015deep}, the Taylor expansion of the first-order loss function~\cite{mozer1988skeletonization, molchanov2019pruning}, the Taylor expansion of the second-order loss function~\cite{lecun1989optimal, molchanov2019importance}, and other variants~\cite{louizos2018learning, yu2018nisp, singh2020woodfisher}.

% Recent research in pruning has been conducted to improve the efficiency of the neural network training stage. This process, known as dynamic sparse training~\cite{mocanu2018scalable,dettmers2019sparse, evci2020rigging}, involves adjusting the pruned model structure throughout the training process while keeping the desired sparsity. However, these approaches simply discard parameters with lower magnitude rankings and overlook the memory footprint of feature map caches, resulting in accuracy drop and inefficient memory savings.

\subsection{Federated Neural Network Pruning}
Federated learning has recently emerged as a promising technique to navigate data privacy challenges in collaborative machine learning~\cite{mcmahan2017communication,zhuang2020fedreid,zhuang2021collaborative,Zhuang22FedEMA}. %, with the FedAvg algorithm~\cite{mcmahan2017communication} leading the charge by emphasizing localized model updates over raw data transfers. 
However, numerous previous federated pruning efforts~\cite{shao2019privacy, li2021lotteryfl, liu2021adaptive, munir2021fedprune} have encountered setbacks because they rely on the training of dense models on devices, which require high computation and memory. Thus, they are not suitable for cross-device FL \cite{kairouz2021advances}, where clients are devices with resource constraints.
% falter due to their reliance on training dense models on devices, misfitting the cross-device FL paradigm, where participants are resource-constrained devices.

%Newer
Recent studies~\cite{qiu2022zerofl,jiang2022model,bibikar2022federated,huang2022fedtiny} introduce on-device pruning via the dynamic sparse training technique~\cite{mocanu2018scalable,dettmers2019sparse, evci2020rigging}. For example, ZeroFL~\cite{qiu2022zerofl} divides the weights into active and non-active weights for inference and sparsified weights and activation for backward propagation. FedDST~\cite{bibikar2022federated} and FedTiny~\cite{huang2022fedtiny}, inspired by RigL~\cite{evci2020rigging}, perform pruning and growing on devices, with the server generating a new global model through sparse aggregation. However, these methods are unable to reduce the memory footprints of the activation cache and suffer from significant accuracy degradation after pruning, since they directly prune parameters that may contain important information.

Therefore, all existing federated neural network pruning works fail in creating a specialized pruned model that can concurrently satisfy both accuracy and memory constraints. To the best of our knowledge, FedMef is the first work that can simultaneously address both issues. 

%Existing federated neural network pruning fails \llv{All the existing federated neural network pruning works fail} in creating a specialized pruned model that meets both accuracy and memory budget requirements. %Therefore, we propose FedMef, which contains budgeted dynamic pruning and scaled feature pruning to address these issues. 
% \llv{Our proposed FedMef can address all these issues}.

\subsection{Activation Cache Compression}
High-resolution activation tensors are a primary memory burden for modern deep neural networks. Gradient checkpoint~\cite{chen2016training, gruslys2016memory, feng2021optimal}, which stores specific layer tensors and recalculates others during the backward pass, offers a memory-saving solution, but at a high computational cost. Alternatively, adaptive precision quantization methods~\cite{chen2021actnn, liu2022gact, wang2023division} compress activation caches through quantization but introduce time overhead from dynamic bit-width adjustments and dequantization. The activation pruning (sparsification) method~\cite{chen2022dropit}, which sparsifies the activation caches, is lighter than other methods, but relies heavily on batch normalization (BN) layers to guarantee that most of the elements in activation are zero or near zero. Relying on BN layers would be problematic to train with small batches and non-independent and identically distributed (non-i.i.d.) data \cite{li2021fedbn, zhuang2023normalization, zhuang2024fedwon}. As a result, current activation pruning methods are unsuitable for resource-constrained devices in FL. %This dependence is problematic in scenarios with small training batches and non-iid data distributions~\cite{li2021fedbn, zhuang2023normalization} in federated learning, making the current activation pruning method unsuitable for resource-constrained devices in federated settings. 
To address these challenges, our proposed FedMef utilizes scaled activation pruning, effectively compressing activation caches without relying on BN layers.

% High-resolution activation tensors are an important factor in the memory requirements of current deep neural networks (DNNs). To reduce the cache of the activation, the gradient checkpoint stores tensors from some layers and recomputes uncached tensors in the backward pass~\cite{chen2016training, gruslys2016memory, feng2021optimal}. However, this method has an additional computational cost for any memory saving. To address this, \cite{chen2021actnn, liu2022gact, wang2023division} explored lossy compression of the activation cache through adaptive precision quantization, which has time overhead from dynamic bit-width allocation and decompression. DropIT~\cite{chen2022dropit} proposed to reduce the storage of activations by activation pruning(sparsification), which is lighter than adaptive precision quantization methods. Unfortunately, DropIT relies heavily on batch normalization layers to guarantee that the majority of elements in activations are small or near zero. However, when the training batch size is small due to memory constraints on the devices, the batch normalization layers become ineffective.
% Furthermore, the non-iid problem in federated learning makes it difficult for batch normalization layers to normalize the elements in activation to near zero~\cite{li2021fedbn, zhuang2023normalization}. As a result, DropIT~\cite{chen2022dropit} is infeasible for resource-constrained devices in federated settings. In contrast, in our proposed FedMef, we use scaled activation pruning to compress activation caches without relying on any batch normalization layer.

\section{Methodology}
% This section introduces FedMef, beginning with the problem setup and then outlining the design principles. \llv{This section first introduces the problem setup, then outlines the design principles of our proposed FedMef.}
This section first introduces the problem setup and then outlines the design principles of our proposed FedMef.
We then introduce two key components in FedMef: budget-aware extrusion and scaled activation pruning.

\subsection{Problem Setup}
In the cross-device FL scenario, numerous resource-constrained devices collaboratively train %a %collective model, avoiding direct data sharing
better models without direct data sharing~\cite{kairouz2021advances}. In this setting, $K$ devices, each with memory and computational constraints, cooperate to train the model with parameters $\theta$. Every device possesses a distinct local dataset, denoted as $\mathcal{D}_k, k \in \{1,2,\dots, K\}$.  The structure of the pruned model is represented using a mask, $m \in \{0, 1\}^{ |\theta| }$, and $\theta \odot m$ denotes the sparse parameters of the pruned model. Our objective is to derive a specialized sparse model with mask $m$, using the local dataset $\mathcal{D}_k$, to optimize prediction accuracy in FL. During training, the sparsity levels of the mask $s_m$ and the activation caches $s_a$ must be higher than the target sparsity ($s_{tm}$ and $s_{ta}$), which is determined by the memory constraints of the devices. 
% \wm{Add a sentence to explain what are $s_{tm}$ and $s_{ta}$} 
Thus, our optimization challenge is to solve the following problem:
% We investigate the classical cross-device federated learning situation, FedAvg~\cite{mcmahan2017communication}, where $K$ devices collaborate to train the model with parameters $\theta$. Each device has its own local dataset $\mathcal{D}_k, k \in \{1,2,\dots, K\}$. All devices have limited memory and computing power\llv{maybe you can clearly clarify we consider cross-device setting}. \llv{maybe you can formally define this as a concrete problem: sparse model or xxx.} We seek to find a specialized mask $m \in \{0, 1\}^{ \theta }$ to achieve the best prediction performance for federated learning. During training, the density $d_m$ of the sparse mask $m$ and the density $d_a$ of the feature map caches $a$ must not exceed the target density $d_t$, which is determined by the limitation of the device memory resources. The goal is to optimize the following problem:
\begin{equation}
\begin{split}
\min_{\bm{\theta}, m}L(\theta, m) := \sum_{k=1}^{K}p_kL_k(\theta, m, \mathcal{D}_k), \\
\textrm{s.t.} \quad s_m \ge s_{tm}, s_a \ge s_{ta}, 
\end{split}
\end{equation}
where $L_k$ is the loss function of the $k$-th device (\textit{e.g.}, cross-entropy loss), and $p_k$ represents the weight of $k$-th device %accorded 
during model aggregation in the server. 
% \cc{what is the loss function? I cannot find it in the whole paper.} 
Before communicating with the server, each device trains its local model for $E$ local epochs.

\subsection{Design Principles}
To ascertain that specialized sparse models can be developed on resource-constrained devices while maintaining privacy, the prevailing trend is to leverage federated dynamic pruning. However, contemporary methods~\cite{jiang2022model, qiu2022zerofl, bibikar2022federated, huang2022fedtiny} face two pressing issues: significant post-pruning accuracy degradation and high activation memory usage. As illustrated in Figure~\ref{fig:framework}, our framework, FedMef, introduces two solutions to these challenges: budget-aware extrusion and scaled activation pruning.

In the FedMef framework, the server starts by distributing a randomly pruned model to the devices. Subsequently, these devices collaboratively engage in training sparse models using scaled activation pruning. During this phase, the activation cache is pruned during the forward pass, effectively optimizing memory utilization, as illustrated in Figure~\ref{fig:act_prune}. After several iterative training rounds, the devices employ the budget-aware extrusion technique to transfer vital information from low-magnitude parameters to others. Subsequently, the server adjusts the model structure through magnitude pruning and gradient-magnitude-based growing. Due to the information transfer facilitated by budget-aware extrusion, the post-pruning accuracy degradation is slight. Finally, the framework continues with the training and adjustment of the sparse model until convergence. 

% To mitigate post-pruning accuracy degradation, budget-aware extrusion facilitates information extrusion from parameters slated for pruning to others, aiming to reduce information loss during pruning. 
Specifically, budget-aware extrusion achieves information transfer by employing a surrogate loss function with $L_1$ regularization of low-magnitude parameters. This process not only suppresses their magnitude but also transfers their information to alternate parameters. Additionally, the devices set up a budget-aware schedule to speed up the extrusion. In the scaled activation pruning, after each layer's forward pass, the activation caches are pruned to reduce memory usage. During the backward pass, the sparse activation caches are used directly. To ensure that the pruned elements are zero or nearly zero, even when training with small batch size, we adopt the Normalized Sparse Convolution (NSConv) to reparameterize the convolution layers instead of using batch normalization layers. Next, we delve into in-depth discussions of budget-aware extrusion and scaled activation pruning techniques.

\begin{figure*}[t]
  \centering
  \includegraphics[width=0.95\linewidth]{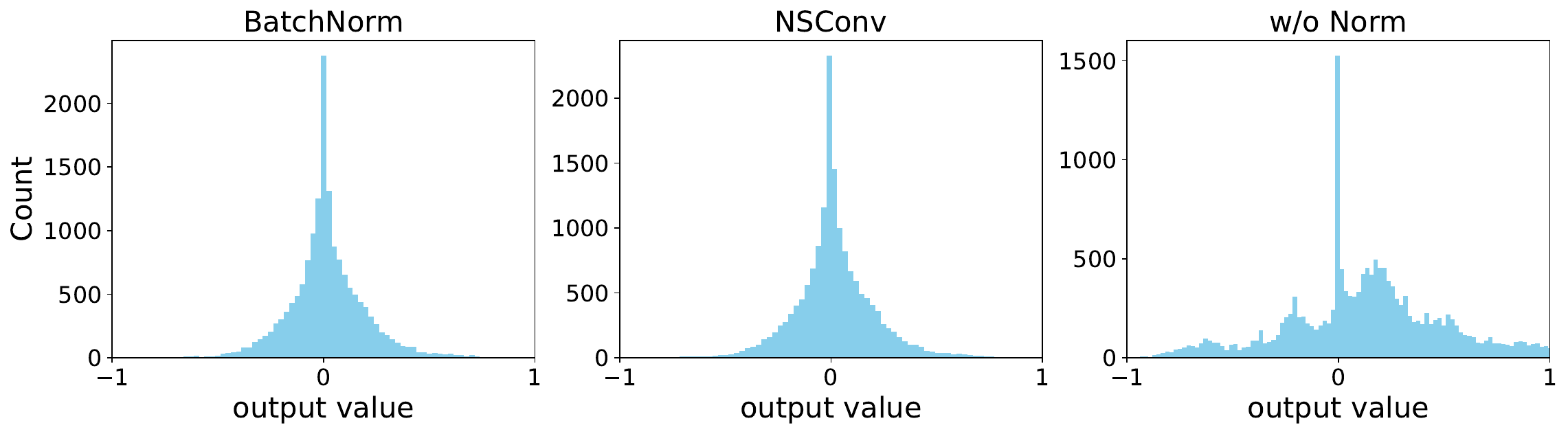}
  \vspace{-0.3cm}
  \caption{Distribution of output from a convolution layer in ResNet18 using batch normalization layers (BatchNorm), without normalization layers (w/o Norm), and with our proposed Normalized Sparse Convolution (NSConv). The output experiences an internal covariate shift when training without normalization layers, whereas NSConv effectively mitigates this issue. 
  % Without the use of normalization layers, the output experiences an internal covariate shift; however, NSConv effectively mitigates this issue. 
  Figure~\ref{fig:hist} in the appendix shows the output distribution for all convolution layers in the ResNet18 model.
  % \wm{The color is a bit too light. Increase the font size of x and y axis, ticks, and titles}
  }
  \vspace{-0.3cm}
  \label{fig:sap_effect}
\end{figure*}

\subsection{Budget-aware Extrusion}
It is essential to address the information loss that occurs during pruning, as the parameters to be pruned often retain valuable information. Direct pruning can cause a substantial accuracy drop,
% \cc{I cannot see the accuracy drop from this figure Maybe we can report the accuracy before and after pruning?} 
demanding considerable resources for recovery, as illustrated in Figure~\ref{fig:framework}. This issue may become even more pronounced in federated contexts due to the heterogeneous data distribution across devices, potentially amplifying the negative impact on model performance during training.

To address this challenge, we take inspiration from the Dual Lottery Ticket Hypothesis (DLTH)~\cite{bai2022dual}. The DLTH suggests that a randomly selected subnetwork can be transformed into one that achieves better, or at least comparable, performance to benchmarks. Building on this premise, we introduce budget-aware extrusion within our FedMef framework, which can extrude the information from the parameters to be pruned to other surviving parameters. After sufficient extrusion, the parameters designated for pruning retain only marginal influence on the network, ensuring minimal information and accuracy loss during pruning.
% the pruning process.

In alignment with the findings of the DLTH~\cite{bai2022dual}, we employ a regularization term to execute this information extrusion. Given the parameters $\theta$ and its associated mask $m$, the extrusion process on the $k$-th device can be realized through the surrogate loss function $L^s_k$:
\begin{equation}
    L^s_k = L_k(\theta, m, \mathcal{D}_k) + \lambda ||\theta_{low}||_2^2,
    \label{eq:loss}
\end{equation}
where $\lambda $ is constant and $\theta_{low}$ represents the parameters earmarked for pruning, which is the subset of unpruned parameters $\theta \odot m$ with the lowest weight magnitudes. 

The inherent constraints associated with edge device training resources require that information extrusion should be executed within a limited budget before the pruning process. However, adhering to the original learning schedule represented by $\eta$ is sub-optimal, as in the later epochs of training, the learning rate following traditional decay mechanisms becomes significantly small, impeding the information extrusion process. To address this issue, we introduce a budget-aware schedule in the context of budget-aware extrusion. \hong{Our proposed budget-aware schedule is constructed to accelerate the extrusion process, especially when the original learning rate is insufficient for rapid extrusion. We have adopted one of the SOTA budgeted schedules, the REX schedule~\cite{chen2022rex}, whose learning rate in the $t$ -th step is represented as $\beta_t = p(t)\eta_0$, where $\eta_0$ represents the initial learning rate in the original learning schedule and $p(t)$ is the REX schedule factor. However, the REX schedule does not take into account the status of the extrusion process, often resulting in an excessively high lr. Therefore, we introduce a scaling term that ranges from 0 to 1, $2\sigma(||\theta_{low}||)-1$ to moderate the budgeted learning rate based on the extrusion status, which is represented by the magnitude of the marked parameters.
Given $T_{budget}$ as the training budget and $t$ as the present step, our proposed budget-aware learning rate $\beta_t $ is mathematically defined as:
\begin{equation}
    \beta_t = p(t)(2\sigma(||\theta_{low}||)-1)\eta_0,
\end{equation}
where the REX schedule factor $p(t)$ is defined by $p(t) = \frac{2T_{budget}-2t}{2T_{budget}-t}$. The main objective behind introducing this factor is to effectively adjust the learning rate based on the relative progression of the training and the preset training budget.  }
During the information extrusion process, the learning rate $\mu_t$ is formulated as follows:
\begin{equation}
\mu_t = \max(\eta_t, \beta_t),
\label{eq:lr}
\end{equation}
where $\eta_t$ is the learning rate in the original schedule. This ensures efficient and timely information extrusion by adjusting an adequate learning rate even in the later stages of training.
% This ensures that even in the later stages of training, the process remains energized by a heightened learning rate, ensuring efficient and timely information extrusion. 
During the normal training stage, the learning rate is $\mu_t = \eta_t$.

In particular, upon receiving the pruned model from the server, the devices mark the parameters $\theta_{low}$ that have the lowest weight magnitude. Then, the devices perform several epochs of budget-aware extrusion with the surrogate loss $L_k^s$ in Equation~\ref{eq:loss}. The learning rate for this phase is dynamic and is governed by the function presented in Equation~\ref{eq:lr}. After the extrusion phase, each device calculates the Top-K gradients across all parameters and uploads the gradients along with the parameters to the server, as indicated in~\cite{huang2022fedtiny}. The server then aggregates the sparse parameters and gradients to obtain the average parameters and average gradients. Finally, the server prunes the marked parameters $\theta_{low}$ and grows the same number of parameters with the largest averaged gradient magnitude.

According to the pruning and growing process, the server creates a global model with a new structure and then FedMef begins to train the new global model. FedMef periodically performs adjustments and training to deliver an optimal sparse neural network suitable for all devices. The pseudo code can be viewed in Algorithm~\ref{alg:alg} in the appendix.

\subsection{Scaled Activation Pruning}
In cross-device FL, where devices may have extremely limited memory, small batch sizes are often employed during training. This diminishes the effectiveness of batch normalization (BN) layers in such a scenario.
% In scenarios where device memory is highly constrained, leading to small training batch sizes, the efficacy of the batch normalization layer diminishes. 
However, current activation cache compression techniques, such as DropIT
% that proposed by Chen et al.
~\cite{chen2022dropit}, are limited in their ability to conduct training without BN layers.
% handle neural networks without a batch normalization layer. 
To address this issue, we propose a scaled activation pruning technique that achieves superior performance even with small batch sizes, as illustrated in Figure~\ref{fig:act_prune}.
% as an innovative solution.

Given a CNN model with ReLU-Conv ordering, in the $l$-th convolution layer, the sparse filters are represented as $\theta^l \in \mathrm{R}^{ks \times ks \times c_{in} \times c_{out}}$, where $ks$ denotes the kernel size; $c_{in}$ and $c_{out}$ denote the number of input and output channels, respectively. For an input value $a^{l-1}$,  the convolution operation in the $l$-th layer that yields the output value $a^{l} $ is:
\begin{equation}
    a^l = \mathrm{Conv}(\theta^l, f(a^{l-1})),
\end{equation}
where $f(\cdot)$ is any activation function such as ReLU~\cite{agarap2018deep}. Note that
% It is worth noting that 
$a^{l-1}$ is not only an input of $l$ layer but also the output of the ${l-1}$-th layer.
During the forward pass, the activation $f(a^{l-1})$ must be retained in memory to compute the gradients of the filters $\theta^l$ during the backward pass. Similarly, for each layer, the activation $f(a^{l})$ must be stored for later usage, which causes substantial memory footprints.

The activation pruning approach, DropIT~\cite{chen2022dropit}, prunes $f(a^{l})$ in the forward pass.
% The approach of Chen et al.~\cite{chen2022dropit} for activation pruning works by pruning $f(a^{l})$ during the forward pass. 
It then uses sparse activation $S(f(a^{l}))$ for gradient computation in the backward pass. This approach requires that the input $a^l$ be centered around zero. This centering ensures a minimized disparity between the sparse activation $S(f(a^{l}))$ and its original counterpart $f(a^{l})$. However, this zero-centered requirement becomes unattainable when the efficacy of the batch normalization layer decreases. This ineffectiveness arises from internal covariate shift issues~\cite{ioffe2015batch, brock2021characterizing}.
% \cc{something wrong with the citation}
Figure~\ref{fig:sap_effect} shows the mean shift in activation within a ResNet18 model without a normalization layer, resulting in a non-zero mean in the activation distribution. The mathematical details of this effect can be found in Appendix~\ref{sec:ics}.

To reduce the disparity between the original and pruned activation, inspired by the recent methods that remove BN layers~\cite{brock2021characterizing, zhuang2023normalization}, we introduce Normalized Sparse Convolution (NSConv) into activation pruning. Our primary objective is to ensure that the output of the convolution layer is consistently centered around zero, \textit{ i.e.,} the mean value is zero. The convolution operation of NSConv at the $l$-th layer is given by:
\begin{equation}
    a^l = \mathrm{Conv}(\hat{\theta}^l, f(a^{l-1})),
    \label{eq:fp}
\end{equation}
where $\hat{\theta}^l$ represents the sparse normalized filters with filter-wise weight standardization. The filter-wise standardization formula of the $i$-th sparse filter, denoted by $\hat{\theta}^l_i \in \mathbb{R}^{ks \times ks \times c_{in}}$, is given by:
\begin{equation}
    \hat{\theta}^l_i = \gamma\sqrt{c_{in}} \frac{\theta^l_i \odot m^l_i - \mu_\theta}{\sigma_\theta},
    \label{eq:WSSConv}
\end{equation}
where $\theta^l_i \in \mathbb{R}^{ks \times ks \times c_{in}} $ specifies the $i$-th filter of the original filters, $\gamma$ is a constant, and $m^l_i$ denotes the corresponding mask for the sparse filter $\theta^l_i$. The terms $\mu_\theta$ and $\sigma_\theta$ represent the mean and standardization value of the sparse filter $\theta^l_i$, excluding the pruned parameters whose corresponding mask is 0. 
% Using NSConv, the following theorem emerges \wm{emerges? We can obtain the following theorem using NSConv?}:
\begin{theorem} Given a CNN model structured in a ReLU-Conv sequence and %allowing the 
$l$-th convolution layer %to 
performing operations as depicted by the forward pass in Equation~\ref{eq:fp} and NSConv in Equation~\ref{eq:WSSConv}, for the $i$-th channel of the activation value, $f(a^{l-1}_i)$, with its mean and variance denoted as $\mu_f, \sigma_f^2$, the mean and variance for the $i$-th channel of the output value, $a^l_i$, will be:
\begin{equation}
\mathbb{E}[a^l_i] = 0, \quad \mathrm{Var}[a^l_i]=\gamma^2(\sigma^2_f+\mu_f^2).
\end{equation}
\label{thm}
\end{theorem}\vspace{-0.5cm}            
Theorem~\ref{thm} reveals insights into the capabilities of scaled activation pruning. Specifically, it highlights its efficacy in addressing the disparity between pruned and original activation in CNNs without BN layers. A key factor in its effectiveness is NSConv's ability to normalize the output of each convolution layer, centering it around zero, as shown in Figure~\ref{fig:sap_effect}.
% A cornerstone of its effectiveness is the NSConv's ability to normalize the output of each convolution layer, making it have a distribution that is centered around zero. 
By adjusting the hyperparameter $\gamma$, we can control the variance of the distribution, causing a large portion of the activation elements to be either zero or close to it. The proof of Theorem~\ref{thm} can be found in Appendix~\ref{sec:proof}.

Incorporating NSConv into scaled activation pruning brings several additional advantages: First, NSConv ignores pruned parameters, focusing solely on the remaining ones. This translates to minimal computational overhead and maintains the sparsity of the normalized parameters. Second, NSConv is suitable for training with small batch sizes because there are no inter-dependencies between batch elements.
% operates without necessitating any interdependencies between batch elements, making it especially adept at handling training with smaller batch sizes. 
Lastly, NSConv ensures uniformity between the training and testing phases.

% In detail, during the forward pass in the training process (including badget-aware extrusion), the sparse weight $\theta^l$ is reparameterized to the filter-wise normalized $\hat{\theta}^l$ before being calculated with the input activation $f(a^{l-1})$ from the preceding layer. Then the input activation is pruned by magnitude, resulting in a sparse activation $P(f(a^{l-1}))$  to free up memory. During the backward pass, sparse activation $P(f(a^{l-1}))$ is used to calculate the gradients $\nabla \theta^l$.             
\section{Evaluation} 
In this section, we dive into an in-depth evaluation of our framework, FedMef. We compare it against SOTA frameworks, demonstrating its effectiveness in various testing conditions. In addition, an ablation study reveals the components that make our proposed framework effective.

\subsection{Experimental Setup}
We assess the effectiveness of FedMef in image recognition tasks using three datasets: CIFAR-10~\cite{krizhevsky2009learning}, CINIC-10~\cite{darlow2018cinic}, and TinyImageNet~\cite{le2015tiny}. Notably, the choice of these datasets is motivated by the imperative of ensuring a fair comparison, given that existing federated dynamic pruning frameworks~\cite{huang2022fedtiny,bibikar2022federated,jiang2022model, qiu2022zerofl} focus on simple datasets. We employ the ResNet18~\cite{he2016deep} and MobileNetV2~\cite{sandler2018mobilenetv2} models for evaluation.
We conduct experiments on a landscape of 100 devices. The datasets are divided into heterogeneous partitions via a Dirichlet distribution characterized by a factor of $\alpha=0.5$. We train the models for $R=500$ federated learning rounds, where each round is composed of $E=10$ local epochs. We set the training batch size as $64$ by default. The target parameters sparsity and target activation sparsity are set to $s_{tm} = 0.9, s_{ta} = 0.9$ by default. %To cement the reliability of our findings, each experiment is conducted
The initial learning rate is set as $\eta_0 = 1$ with an exponential decay rate of 0.95.
We conduct each experiment five times and report the average result and standard deviation. 

% We choose FL-PQSU~\cite{xu2021accelerating}, FedDST~\cite{bibikar2022federated}, and FedTiny~\cite{huang2022fedtiny}, as the baseline frameworks. \hh{We selected FL-PQSU to represent static pruning, establishing it as our lower bound. Additionally, we opt for FedDST and FedTiny, as they are the state-of-the-art methods in federated dynamic pruning.} 
We compare our proposed FedMef with FL-PQSU~\cite{xu2021accelerating}, FedDST~\cite{bibikar2022federated}, and FedTiny~\cite{huang2022fedtiny}. FL-PQSU is a static pruning method, which employs an initialized pruning based on the $L_1$ norm on the server prior to training. It can be considered as the lower bound of our method. FedDST and FedTiny are SOTA federated dynamic pruning frameworks. Both of them start with an initial pruned model, subsequently employing mask adjustments to adjust the model architecture.
% FL-PQSU employs an initialized pruning based on the $L_1$ norm on the server prior to training. 
% Both FedDST and FedTiny start off with an initial pruned model, subsequently employing mask adjustments to adjust the model architecture.  %quintessential
The key distinction among these methods lies in their locus of model structure adjustments. FedTiny centralizes this on the server, whereas FedDST decentralizes it to the devices. Certain federated pruning frameworks, such as ZeroFL~\cite{qiu2022zerofl} and PruneFL~\cite{jiang2022model}, which are memory-intensive to process dense models, are consciously excluded from our comparison.

In the FedDST, FedTiny, and FedMef frameworks, the adjustment of the model structure is applied after $\Delta R = 10$ training rounds. Upon reaching $R_{stop} = 300$ rounds, the framework suspends further adjustment, continuing its training until reaching $R=500$ rounds. The pruning number for each layer is set to $0.2(1 + cos\frac{t\pi}{R_{stop}E})n$ in the $t$ -th iteration, where $n$ is the number of unpruned parameters in the $l$-th layer. Due to FedDST~\cite{bibikar2022federated} necessitating a series of on-device training epochs for fine-tuning after adjustment, after 5 epochs of local training, we let FedDST adjust the model structure and then proceed with 5 training epochs. FedTiny's~\cite{huang2022fedtiny} adaptive batch normalization module is amputated from our experiments, as its memory overhead renders it infeasible for our device constraints.

\begin{figure*}[t]
  \centering
  \includegraphics[width=0.95\linewidth]{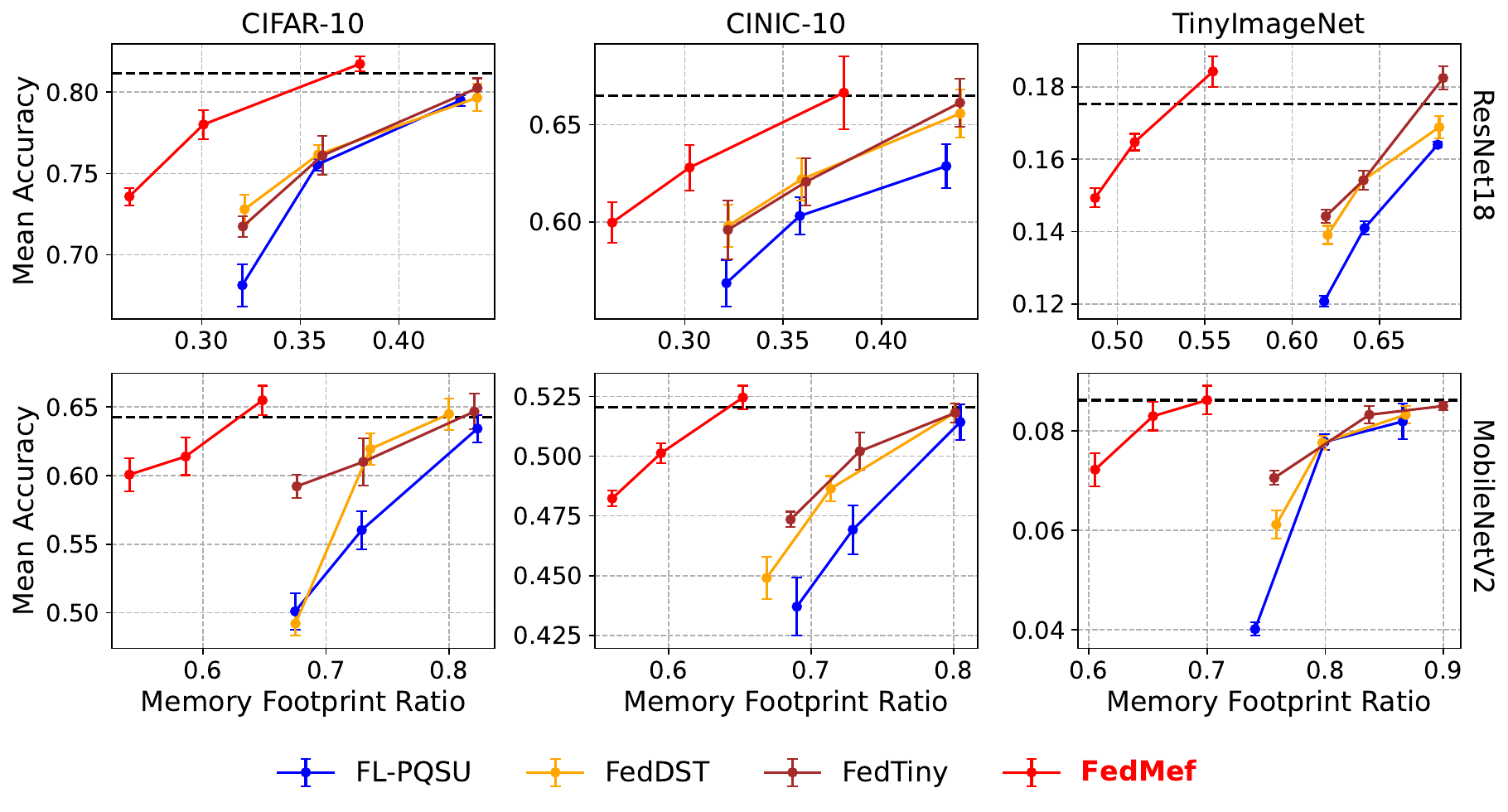}
  \vspace{-0.3cm}
  \caption{Comparison of accuracy and memory footprint of our proposed FedMef with the existing federated pruning methods on three datasets. The black dashed line marks the accuracy of training a full-size model (without pruning) in FedAvg. The memory footprint ratio is the memory footprint relative to training a full-size model in FedAvg.
  % Comparison of FedMef's mean accuracy and standard deviation against other federated pruning frameworks. 
  % The black dashed line marks the accuracy \st{benchmark} of the full-size FedAvg framework. The "Memory Footprint Ratio" indicates each framework's memory footprint relative to the full-size FedAvg training.
  }
  \label{fig:main_result}
  % \vspace{-0.3cm}
\end{figure*}

\subsection{Performance Evaluation}
To demonstrate the effectiveness of FedMef, we compare it with other frameworks on the CIFAR-10, CINIC-10, and TinyImageNet datasets using ResNet18 and MobileNetV2. A holistic comparison is illustrated in Figure~\ref{fig:main_result}.
The target sparsity of the parameters is set to $s_{tm} \in \{ 0.95, 0.9, 0.8\}$. Remarkably, FedMef outperforms all baseline frameworks in terms of accuracy and memory efficiency. %As a case in point,
For instance, FedMef achieves an accuracy improvement of $2.13\%$ on the CIFAR-10 dataset with the MobileNetV2 model, while saving $28.5\%$ memory usage compared to the best baseline framework, FedTiny. Such advances in accuracy can be attributed to budget-aware extrusion, while scaled activation pruning primarily augments memory conservation. %Empirical observations substantiate the potency of both innovations.

An obvious trend is the superior accuracy benchmarks set by ResNet18 over MobileNetV2 in all datasets. The design of MobileNetV2 is tailored to large-scale image classification, which may make it less suitable for relatively small datasets. A noteworthy observation is that FedTiny generally outperforms other baseline methods within comparable memory footprints. Given this empirical trend, ResNet18 is chosen as the default model and FedTiny serves as the primary reference for subsequent experiments. 

\textbf{Computational and Communication Cost.}
While FedMef exhibits superior performance compared to various baseline frameworks with relatively low memory footprints, it is imperative to conduct a comprehensive analysis of the computational and communication costs inherent in the FedMef framework.
To illustrate the computational and communication costs, we evaluated FedMef on the CIFAR-10 dataset using the ResNet18 model. The results, presented in Table~\ref{tb:cost}, elucidate that FedMef induces marginal communication and computational overhead. For example, with a target density of $s_{tm} = 90\%$, FedMef introduces only a mere computational overhead of $0.003\times$ and a communication overhead of $0.005\times$, while improving the accuracy by $2\%$ compared to other baselines. The detailed analysis for the training cost is shown in Appendix~\ref{app:compression}.

% \usepackage{multirow}
\begin{table}[]
\begin{tabular}{ccccc}
\hline
$s_{tm}$& Framework & \begin{tabular}[c]{@{}c@{}}Mean \\ Accuracy\end{tabular} & \begin{tabular}[c]{@{}c@{}}Training \\ FLOPs\end{tabular} & \begin{tabular}[c]{@{}c@{}}Data \\ Exchange\end{tabular} \\
\hline 
0 & FedAVG & 81.2\% & \begin{tabular}[c]{@{}c@{}}1$\times$\\      (8.33e12)\end{tabular} & \begin{tabular}[c]{@{}c@{}}1$\times$\\      (89.52MB)\end{tabular} \\
\hline
\multirow{3}{*}{95\%} & FedDST & 72.8\% & 0.057$\times$ & 0.083$\times$ \\
 & FedTiny & 71.8\% & 0.057$\times$ & 0.086$\times$ \\
 & \textbf{FedMef} & \textcolor{Red}{\textbf{73.6\%}} & 0.061$\times$ & 0.086$\times$ \\
 \hline
\multirow{3}{*}{90\%} & FedDST & 76.2\% & 0.113$\times$ & 0.133$\times$ \\
 & FedTiny & 76.1\% & 0.113$\times$ & 0.138$\times$ \\
 & \textbf{FedMef} & \textcolor{Red}{\textbf{78.1\%}} & 0.116$\times$ & 0.138$\times$ \\
 \hline
\multirow{3}{*}{80\%} & FedDST & 79.7\% & 0.218$\times$ & 0.232$\times$ \\
 & FedTiny & 80.3\% & 0.218$\times$ & 0.243$\times$ \\
 & \textbf{FedMef} & \textcolor{Red}{\textbf{81.7\%}}& 0.221$\times$ & 0.243$\times$ \\
 \hline
\end{tabular}
\caption{Accuracy and training cost of proposed FedMef and other baseline framework. We report the maximum FLOPs for training (Training FLOPs ) and the maximum data exchange for communication (Data Exchange). All cost measurements are for one device in one adjustment round.}
\label{tb:cost}
\vspace{-0.3cm}
\end{table}

\begin{figure*}[!t]
  \centering
  \includegraphics[scale=0.38]{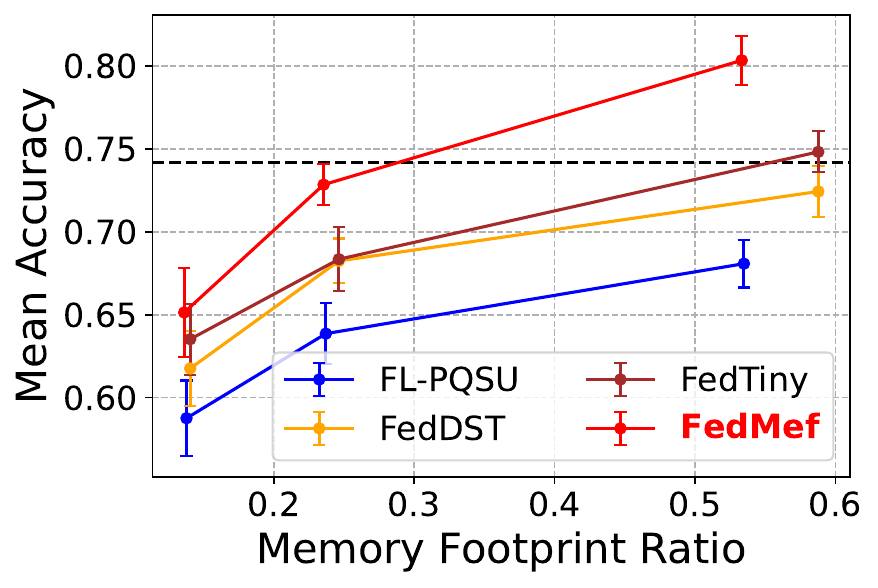}
  \hspace{0.cm}
  \includegraphics[scale=0.38]{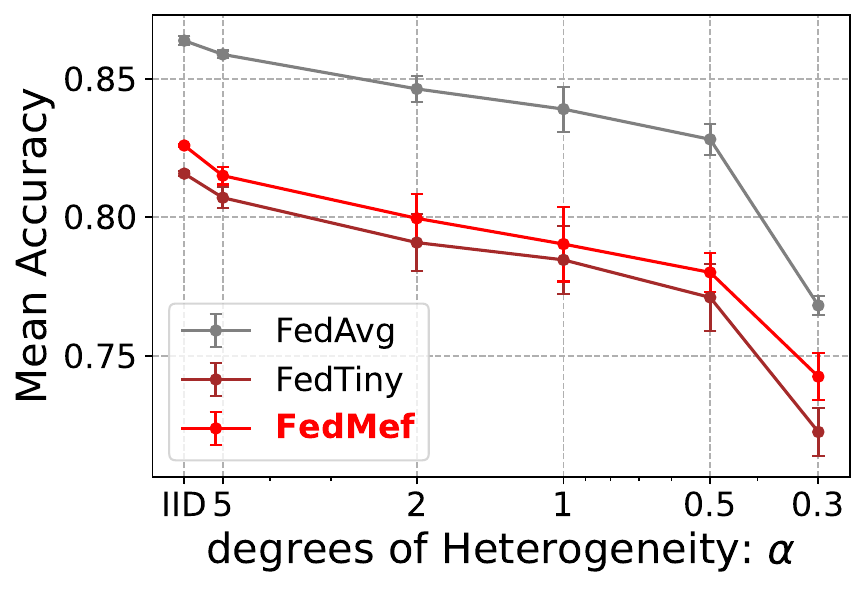}
  \hspace{0.cm}
  \includegraphics[scale=0.38]{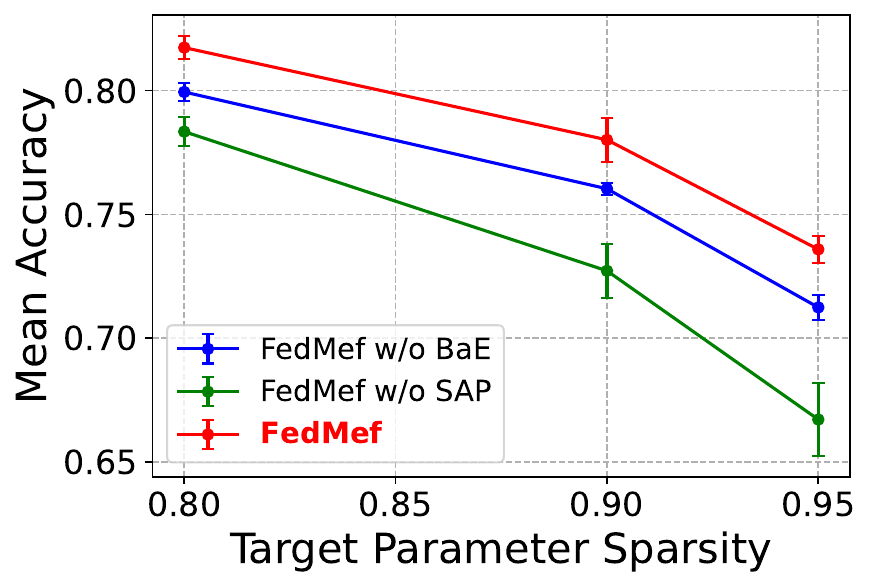}
  \vspace{-0.3cm} 
  \caption{FedMef's average accuracy and standard deviation are compared against: \textit{(left)} various federated pruning frameworks when the training batch size is 1, where the black dashed line represents the accuracy of FedAvg framework; \textit{(middle)} FedAvg and FedTiny across varying degrees of data heterogeneity; \textit{(right)} modified versions of FedMef - one excluding BaE (similar to FedTiny's approach) and the other without SAP (omitting NSConv).}
  % \vspace{-0.3cm}
  \label{fig:other_1}
\end{figure*}

\textbf{Training with Small Batch Size.}
Under strict memory constraints, training requires a smaller batch size. However, this compromises statistical robustness and often hinders the effectiveness of batch normalization. To address this issue, we propose scaled activation pruning. The evaluations conducted on the CIFAR-10 dataset with ResNet18, where the batch size is set to 1, as shown in Figure~\ref{fig:other_1} (left), demonstrate that FedMef outperforms all baseline methodologies. The significant improvement in accuracy is mainly due to the use of NSConv in scaled activation pruning, which is further demonstrated in the appendix~\ref{sec:wssconv}.

\textbf{The Impact of Adjustment Period.} After model structure adjustment, it is necessary to restore accuracy loss through several training rounds. Therefore, the adjustment period should be longer. Unfortunately, given the computational constraints of certain devices, there is an urgent need to limit the number of interval training rounds and local epochs. The empirical results of the experiments on the CIFAR-10 dataset with various adjustment periods, $\Delta R$, and a single local epoch are shown in Table~\ref{tb:R}. When FedTiny performance decreases under resource constraints, FedMef remarkably maintains the performance. %exhibits remarkable resilience.

\begin{table}[t]
    \centering
    \begin{tabular}{c|cc}
    \hline
    $\Delta R$ & FedTiny & FedMef \\ \hline
      3  & 55.09\%(1.82\%)  &   61.94\%(0.49\%)  \\
     5 &  58.73\%(1.62\%) &  62.12\%(0.58\%) \\
     10 & 61.18\%(1.08\%) & 62.77\%(0.78\%) \\
    \hline
    \end{tabular}
    \caption{ Mean accuracy (standard deviation) for FedMef and FedTiny on CIFAR-10 with various adjustment periods, $\Delta R$.}
    \label{tb:R}
    \vspace{-0.3cm}
\end{table}

\begin{figure}[t]
  \centering
  \includegraphics[width=0.7\linewidth]{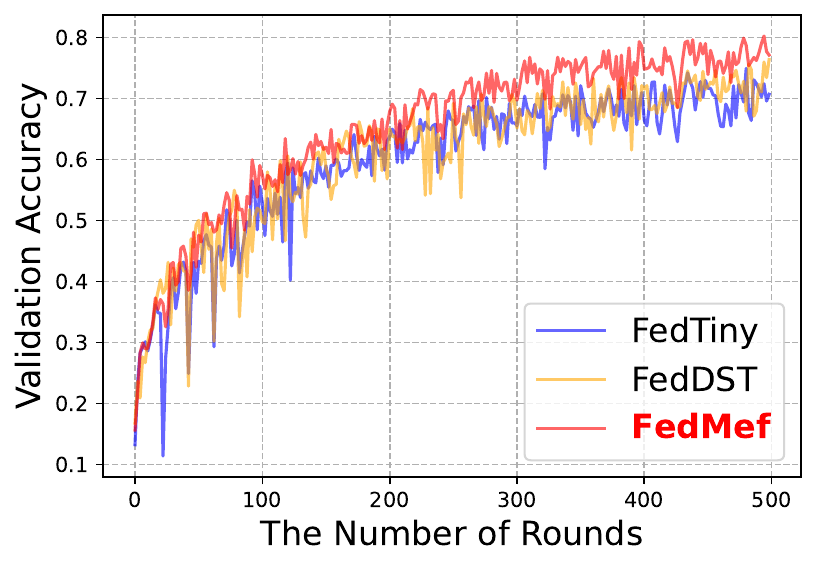}
  \vspace{-0.3cm}
  \caption{The validation accuracy during the training in FedMef and the baseline frameworks.}
  \label{fig:converge}
  \vspace{-0.3cm}
\end{figure}

\textbf{Analysis on Convergence Behavior.} Given FedMef's capability to preserve post-pruning performance through budget-aware extrusion, we anticipate that its convergence speed will surpass that of other baseline frameworks, particularly with a smaller adjustment period. To assess this, we evaluate the performance of FedMef and other baseline frameworks in the CIFAR-10 datasets with the adjustment period ($\Delta R$) set to 5. The results illustrate that the convergence trajectory of FedMef is notably more stable than other baselines, reaching a higher final accuracy, as shown in Figure~\ref{fig:converge}. This observation underscores the efficacy of BaE in enhancing the convergence behavior of FedMef.

\hong{\textbf{Analysis on Different Degrees of Data Heterogeneity.}} We test the effectiveness of FedMef on heterogeneous data distributions by modulating the Dirichlet distribution factor $\alpha$, where the lower $\alpha$ indicates a higher degree of heterogeneity. For reference, we compare our results with the full-size model and FedTiny on the CIFAR-10 dataset and the results are shown in Figure~\ref{fig:other_1} (middle). %Unsurprisingly, 
FedMef retains its superior performance compared to the SOTA framework.

% \begin{wrapfigure}{l}{0.5\textwidth}
% \includegraphics[width=0.9\linewidth]{Image/nonIID.pdf} \caption{The accuracy of various pruning techniques with various degrees of non-iid. The lower the $\alpha$ value, the more non-iid the datasets become.} 
% \label{fig:noniid}
% \end{wrapfigure}

% \vspace{-1cm}

\subsection{Ablation Study}
We further analyze the individual contributions of budget-aware extrusion and scaled activation pruning using trials on the CIFAR-10 dataset with ResNet18. The variants include a FedMef without budget-aware extrusion (akin to FedTiny's mechanism) and a FedMef without scaled activation pruning (mirroring DropIT's approach~\cite{chen2022dropit} without NSConv). The findings presented in Figure~\ref{fig:other_1} (right) indicate that both budget-aware extrusion and scaled activation pruning boost FedMef's performance. In particular, removing scaling in activation pruning results in substantial information loss during backpropagation and leads to performance degradation.

% \begin{wrapfigure}{r}{0.5\textwidth}
% \includegraphics[width=0.9\linewidth]{Image/ablation.pdf} \caption{Ablation study highlighting the two pivotal techniques of FedMef: budget-aware extrusion(BaE) and scaled activation pruning (SAP). We compare FedMef, FedMef without BaE (akin to FedTiny's mechanism), FedMef without SAP (mirroring DropIT's approach). } 
% \label{fig:ablation}
% \end{wrapfigure}

% \begin{wrapfigure}{l}{0.5\textwidth}
% \includegraphics[width=0.9\linewidth]{Image/ablation.pdf} \caption{Ablation studies the two key techniques in FedMef: the budgeted parameter pruning and the scaled feature pruning. We compare FedMef, FedMef without budgeted parameter pruning, FedMef without scaled feature pruning. We tested the ResNet18 model on the CIFAR-10 dataset with various densities.} 
% \label{fig:ablation}
% \end{wrapfigure}

% \begin{figure*}[t]
%   \centering
%   \includegraphics[width=0.5\linewidth]{Image/ablation.pdf}
%   \caption{Ablation studies the two key techniques in FedMef: the budgeted parameter pruning and the scaled feature pruning. We compare FedMef, FedMef without budgeted parameter pruning, FedMef without scaled feature pruning. We tested the ResNet18 model on the CIFAR-10 dataset with various densities.}
%   \label{fig:ablation}
% \end{figure*}

% batch-size = 1, 64f,  | 64 w/o bn?  
% experiment. 
% batch-size = 1, 64f | 64 ?

% 200/10, 100/10, 20/10, num of client
\section{Conclusion}
This paper introduces FedMef, a memory-efficient federated dynamic pruning framework designed to generate specialized models on resource-constrained devices in cross-device FL. FedMef addresses the issues of post-pruning accuracy degradation and high activation memory usage that current federated pruning methods suffer from. It proposes two new components: budget-aware extrusion and scaled activation pruning. Budget-aware extrusion reduces information loss in pruning by extruding information from parameters marked for pruning to other parameters within a limited budget. Scaled activation pruning allows activation caches to be pruned to save more memory footprints without compromising accuracy. Experimental results demonstrate that FedMef outperforms existing approaches in terms of both accuracy and memory footprint. %FedMef reduces the memory footprint by 28.5\% compared to the most state-of-the-art method while improving the accuracy by more than 2\%
\newpage
{
    \small
    \bibliographystyle{ieeenat_fullname}
    \bibliography{main}
}

\clearpage
\setcounter{page}{1}
\maketitlesupplementary

\begin{algorithm*}
\caption{FedMef}
\label{alg:alg}
\textbf{Input}: dense initialized parameters $\theta$, $K$ devices with local dataset $\mathcal{D}_1, \dots \mathcal{D}_K$, iteration number $t$, learning rate schedule $\alpha_t$, adjustment schedule $\xi^l_t$, denoting the number of adjustment parameters for each layer $l$, the number of local epochs per round $E$, the number of rounds between two adjustment $\Delta R$, and the rounds at which to stop adjustment $R_{stop}$.\\
\textbf{Output}: a well-trained model with sparse $\theta_{t}$ and specified mask $m_{t}$\\
\begin{algorithmic}[1] %[1] enables line numbers
\STATE $t \leftarrow 0$
\STATE $\theta_0, m_0 \leftarrow$ random prune dense initialized parameters $\theta$
\WHILE{until converge}
\FOR{each device $k = 1$ to $K$}
\STATE Fetch sparse parameters $\theta_t$ and mask $m_t$ from the server
\FOR{$i = 0$ to $E-1$}
\STATE $\hat{\theta}^k_{t+i} \leftarrow $ Filter-wise Sparse Standardization as in Equation~\ref{eq:WSSConv}.
\IF{$t \mod \Delta RE = 0$ and $t \leq ER_{stop}$}
\STATE Calculate budget-aware learning rate $\beta_{t+i}$ as in Equation~\ref{eq:lr}.
\STATE $\mu_{t+i} \leftarrow \mathrm{max}(\eta_{t+i}, \beta_{t+i})$
\STATE $\theta^k_{t+i+1} \leftarrow \theta^k_{t+i} - \mu_{t+i}\nabla L^s_k(\hat{\theta}^k_{t+i}, m_{t}, \mathcal{D}_{t+i}^k) \odot m_{t}$, using scaled activation pruning
\ELSE
\STATE $\theta^k_{t+i+1} \leftarrow \theta^k_{t+i} - \eta_{t+i}\nabla L_k(\hat{\theta}^k_{t+i}, m_{t}, \mathcal{D}_{t+i}^k) \odot m_{t}$, using scaled activation pruning
\ENDIF
\ENDFOR
\STATE Upload $\theta^k_{t+E}$ to the server
\IF{$t \mod \Delta RE = 0$ and $t \leq ER_{stop}$}
\FOR{each layer $l$ in model}
\STATE Compute top-$\xi^l_t$ gradients $\tilde{\bm{g}}^{k,l}_t$ for pruned parameters with a memory space of $O(\xi^l_t)$
\STATE Upload $\tilde{\bm{g}}^{k,l}_t$ to the server
\ENDFOR
\ENDIF
\ENDFOR
\STATE 
\STATE The server does
\STATE $\theta_{t+E} \leftarrow \sum_{k=1}^K\frac{|\mathcal{D}_k|}{\sum_{k=1}^{K}|\mathcal{D}_k|}\theta^k_{t+E}$
\IF{$t \mod \Delta RE = 0$ and $t \leq ER_{stop}$}
\FOR{each layer $l$ in model}
\STATE $\tilde{\bm{g}}^l_t \leftarrow \sum_{k=1}^{K}\frac{|\mathcal{D}_k|}{\sum_{k=1}^{K}|\mathcal{D}_k|} \tilde{g}^{k,l}_t$
\STATE $\bm{I}^l_{grow} \leftarrow$ the $\xi^l_t$ pruned indices with the largest absolute value in $\tilde{\bm{g}}^l_t$
\STATE $\bm{I}^l_{drop} \leftarrow$ the $\xi^l_t$ unpruned indices with smallest weight magnitude in $\theta_{t+E}$
\STATE Compute the new mask $m_{t+E}^l$ by adjusting $m_{t}^l$ based on $\bm{I}^l_{grow}$ and $\bm{I}^l_{drop}$
\ENDFOR
\STATE $\theta_{t+E} \leftarrow \theta_{t+E} \odot m_{t+E}$ // Prune the model using the updated mask
\ELSE
\STATE$m_{t+E} \leftarrow m_t$
\ENDIF 
\STATE $t \leftarrow t + E$
\ENDWHILE
\end{algorithmic}
\end{algorithm*}

\section{Proof of Theorem 1}
In this section, we first introduce the internal covariate shift in CNN without batch normalization layers and then provide the proof of Theorem 1.

\begin{figure*}[t]
  \centering
  \includegraphics[width=\linewidth]{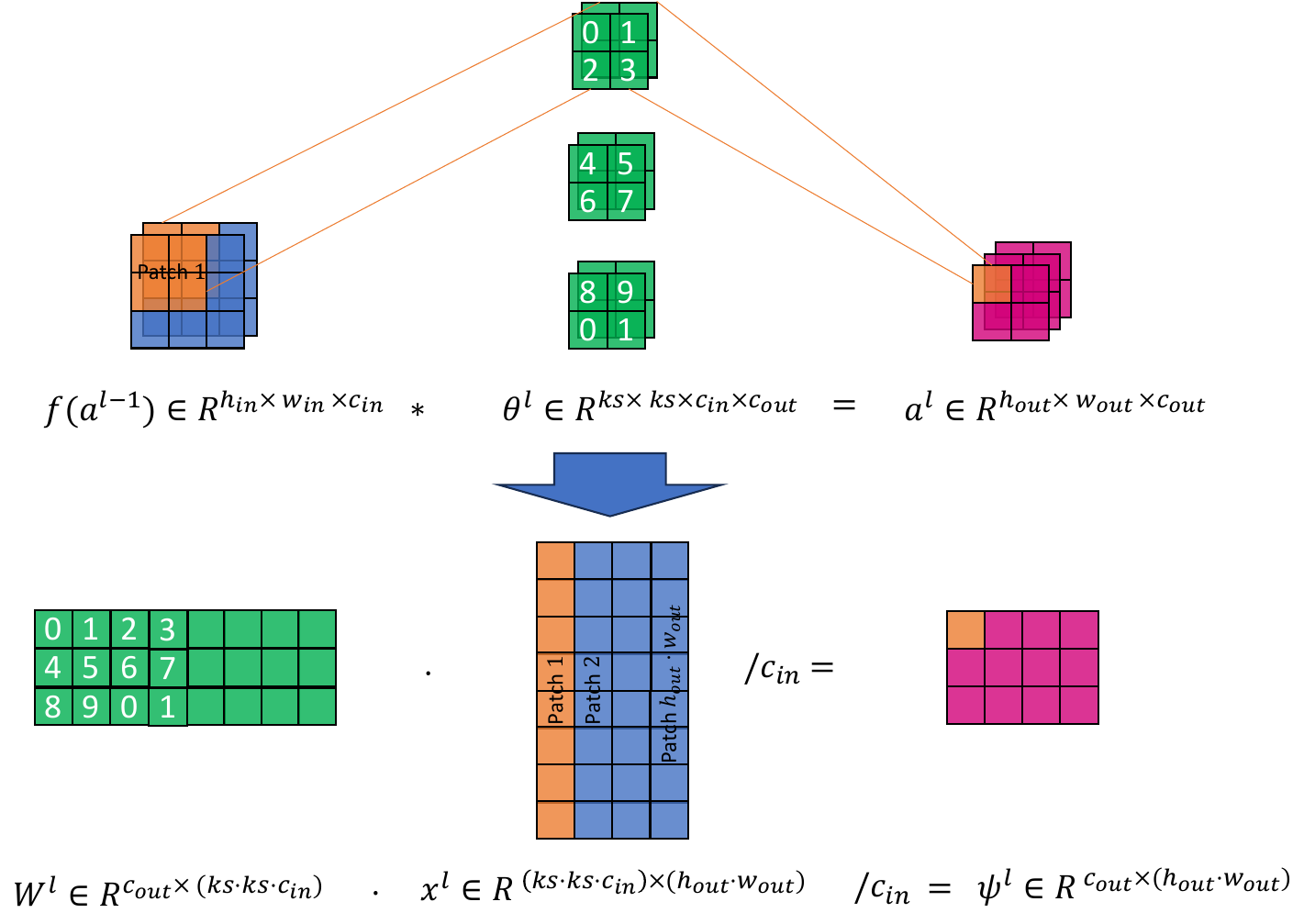}
  \caption{Illustration of transforming the convolution operation into linear multiplication: Start by flattening each filter from the convolutional filters, $\theta^l$, and stacking them to produce the linear weight $W^l$. Next, stack each convolution patch from the input value $f(a^{l-1})$ to form the linear input $x^l$. The resultant multiplication, $\psi^l$, corresponds to a reshaped version of the original output $a^l$.
  }
  \label{fig:transform}
\end{figure*}

\begin{figure*}[t]
  \centering
  \includegraphics[width=\linewidth]{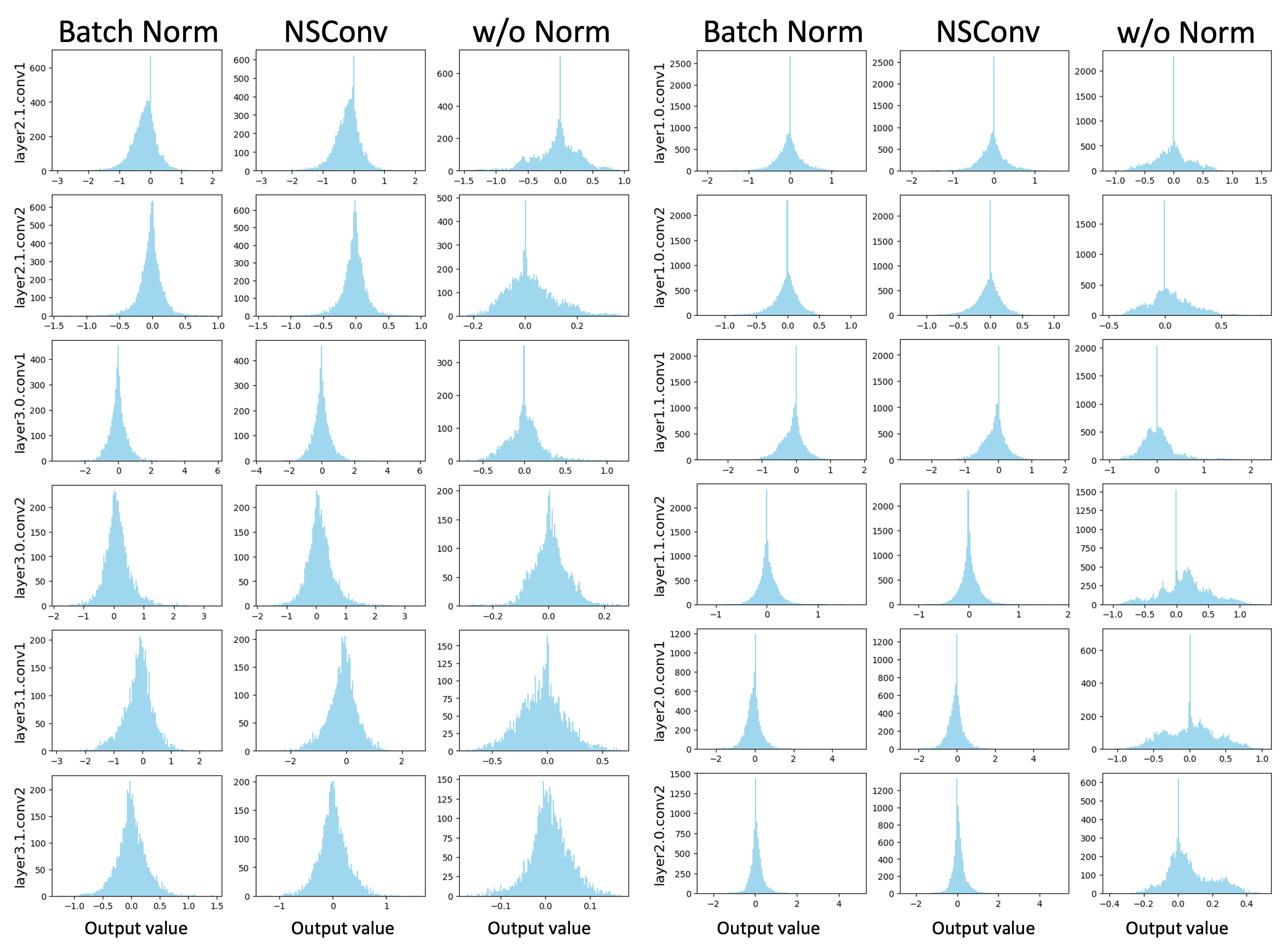}
  \caption{Distribution of output from all convolution layers in ResNet18 model using Batch Normalization
layers (BatchNorm), without normalization layers (w/o Norm), and with Normalized Sparse Convolution (NSConv). 
The range of activation values exhibits a decrease, and the distribution becomes more centralized as the layer deepens. This observation aligns with Equations~\ref{eq:mean} and~\ref{eq:var}, which suggest that the mean and variance values will be scaled with $1/c_{in}$ and $1/c_{in}^2$, respectively.}
  \label{fig:hist}
\end{figure*}

\subsection{Internal Covariate Shift}
\label{sec:ics}
Given a CNN model with ReLU-Conv ordering, in the $l$-th convolution layer, the sparse filters are represented as $\theta^l \in \mathbb{R}^{ks \times ks \times c_{in} \times c_{out}}$, where $ks$ denotes the kernel size;$c_{in}$ and $c_{out}$ denote the number of input and output channels, respectively. For an input value $a^{l-1} \in \mathbb{R}^{h_{in} \times w_{in} \times c_{in}}$,  the convolution operation in the $l$-th layer that yields the output value $a^{l} \in \mathbb{R}^{h_{out} \times w_{out} \times c_{out}}$ is:
\begin{equation}
    a^l = \mathrm{Conv}(\theta^l, f(a^{l-1})),
\end{equation}
where $f(\cdot)$ is any activation function such as ReLU, leaky ReLU, etc. It should be noted that $a^{l-1}$ is not just an input; it is also the output of the ${l-1}$-th layer.

As illustrated in Figure~\ref{fig:transform}, tabove convolution operation can be converted into a linear multiplicity version as :

\begin{equation}
    \psi^l = W^l x^l/c_{in},
\end{equation}

where weight matrix $W^l \in \mathbb{R}^{c_{out}\times (ks\cdot ks\cdot c_{in})}$ is the flattening version of the convolution filters $\theta^l$. The $i$-th row of the linear weight $W^l$ is the flattening result of the $i$-th filter of the original filters, $\theta^l_i$.
The linear input $x^l \in \mathbb{R}^{(ks\cdot ks\cdot c_{in})\times (h_{out}\cdot w_{out})}$ is the stacked convolution patch from activation $f(a^{l-1})$. The resultant multiplication, $\psi^l$, corresponds to a reshaped version of the original output $a^l$. The $i$-th row of the linear result $\psi^l$ is the flattening result of the $i$-th channel of the original output, $a^l_i$. 

Denote the mean and variance values of the $i$ -th filter of the original filters as $\mathbb{E}(\theta^l_i) = \mu_\theta$ and $\mathrm{Var}(\theta^l_i) = \sigma_\theta^2$. Assuming the mean and variance values of the linear input $x^l$ are $\mathbb{E}(x^l) = \mu_x$ and $\mathrm{Var}(x^l) = \sigma_x^2$, the mean and variance of the $i$-th channel of output $a^l_i$ will be :

\begin{equation}
    \mathbb{E}(a^l_i) =\mathbb{E}(\psi^l_i) = \mathbb{E}(W^l_i)\mathbb{E}(x^l)/c_{in} = \mu^{\theta}_i \mu_x/c_{in},
    \label{eq:mean}
\end{equation}

\begin{equation}
\begin{split}
    \mathrm{Var}(a^l_i) =\mathrm{Var}(\psi^l_i) = \mathrm{Var}(W^l_ix^l)/c^2_{in} \\
      = (\sigma_\theta^2\sigma_x^2+ \sigma_\theta^2\mu_x^2 + \mu_\theta^2\sigma_x^2)/c^2_{in},
\end{split}
\label{eq:var}
\end{equation}

Consider $f(\cdot)$ to be the activation function of ReLU, which implies that the input value $\mu_x$ has a positive mean. During training, the mean value of each filter $\theta^l_i$ is difficult to keep at zero. Therefore, without a batch normalization layer, the mean output from the convolution layer will not reach around zero. 

\subsection{Proof}
\label{sec:proof}

\textbf{Theorem} \textit{Given a CNN model structured in a ReLU-Conv sequence, and allowing the $l$-th convolution layer to perform operations as depicted by the forward pass in Equation~\ref{eq:fp} and NSConv in Equation~\ref{eq:WSSConv}. For the $i$-th channel of the activation value, $f(a^{l-1}_i)$, with its mean and variance denoted as $\mu_f, \sigma_f^2$. The mean and variance for the $i$-th channel of the output value, $a^l_i$, will be:
\begin{equation}
\mathbb{E}[a^l_i] = 0, \quad \mathrm{Var}[a^l_i]=\gamma^2(\sigma^2_f+\mu_f^2).
\end{equation}}

\textit{Proof.} As illustrated in Figure~\ref{fig:transform}, convolution operation can be converted to a linear multiplicity version as:

\begin{equation}
    \psi^l = \hat{W}^l x^l/c_{in},
\end{equation}

where the weight matrix $\hat{W}^l \in \mathbb{R}^{c_{out}\times (ks\cdot ks\cdot c_{in})}$ is the flattening version of the sparse normalized convolution filters $\hat{\theta}^l$. The $i$-th row of the linear sparse weight $\hat{W}^l_i$ is the flattening result of the $i$ -th filter of normalized filters,  $\hat{\theta}^l_i$. 

Therefore, the mean and variance of the $i$-th row of normalized linear weight, $\hat{W}^l_i$ are $\mathbb{E}(\hat{W}^l_i) = 0$ and $\mathrm{Var}(\hat{W}^l_i) = \gamma^2c_{in}$.  The mean and variance for the $i$-th of the output value will be:

\begin{equation}
    \mathbb{E}(a^l_i) =\mathbb{E}(\psi^l_i) = \mathbb{E}(\hat{W}^l_i)\mathbb{E}(x^l)/c_{in} = 0, 
\end{equation}

\begin{equation}
\begin{split}
    \mathrm{Var}(a^l_i) =\mathrm{Var}(\psi^l_i) = \mathrm{Var}(\hat{W}^l_ix^l)/c^2_{in} \\
    = \gamma^2(\sigma_x^2+ \mu_x^2),
\end{split}
\end{equation}

Because the linear input $x^l$ is the sampled version of the input activation $f(a^{l-1})$, considering randomness, the mean and variance of the linear input $x^l$ will be $\mu_x = \mu_f$, $\sigma_x^2 = \sigma_f^2$. Therefore, we can get:
\begin{equation}
    \mathbb{E}(a^l_i) = 0, \quad \mathrm{Var}(a^l_i) = \gamma^2(\sigma^2_f+\mu_f^2).
\end{equation}

\subsection{Experiment Result}
\label{sec:wssconv}
To assess the effectiveness of our proposed Normalized Sparse Convolution (NSConv), we conducted experiments on the CIFAR-10 dataset with the ResNet18 model in our proposed FedMef framework, with the sparsity of target parameters set to $0.9$. The results of the experiment, shown in Figure~\ref{fig:hist}, demonstrate that NSConv can achieve an effect similar to that of a Batch Normalization layer. Furthermore, the activation values of ResNet18 without normalization decrease and the distribution becomes more centralized as the layer deepens, further supporting Equations~\ref{eq:mean} and~\ref{eq:var}, which indicate that the mean and variance values will be scaled with $1/c_{in}$ and $1/c_{in}^2$, respectively.

\section{Memory, FLOPs and Communication Costs}
\label{app:compression}
In our experiments, we conducted a comparative analysis between the proposed FedMef and other baselines, focusing on training memory footprints, maximum training FLOPs per round, and communication costs per round. In this section, we first introduce the sparse compression strategies and then present the estimated calculations of the above metrics.

\subsection{Compression Schemes} 
The storage for a matrix consists of two components, values and positions. Compression aims to reduce the storage of the positions of non-zero values in the matrix. Suppose we want to store the positions of $m$ non-zeros value with $b$ bit-width in a sparse matrix $M$. The matrix $M$ has $n$ elements and a shape $n_r \times n_c$. Depending on the density $d = m/n$, we apply different schemes to represent the matrix $M$. We use $o$ bits to represent the positions of $m$ nonzero values and denote the overall storage as $s$.
\begin{itemize}
    \item For density $d \in [0.9, 1]$, \textbf{dense} scheme is applied, i.e. $s = n \cdot b$.
    \item For density $d \in [0.3, 0.9)$, \textbf{bitmap} (BM) is applied, which stores a map with $n$ bits, \textit{i.e.}, $o = n, s = o + mb$.
    \item For density $d \in [0.1, 0.3)$, we apply \textbf{coordinate offset} (COO), which stores elements with its absolute offset and it requires $o = m  \lceil\log_2n \rceil$ extra bits to store position. Therefore, the overall storage is $s = o + mb$
    \item For density $d \in [0., 0.1)$, we apply \textbf{compressed sparse row} (CSR) and \textbf{compressed sparse column} (CSC) depending on size. It uses the column and row index to store the position of elements, and $o = m\lceil \log_2 n_c \rceil + n_r\lceil \log_2m\rceil$ bits are needed for CSR. The overall storage is $s = o + mb$
\end{itemize}

For tenor, we carry out reshaping before compression. This approach allows us to determine the memory needed to train the network's parameters.

\subsection{The Memory Footprint of Training Models} 
We estimate the memory footprint for training to be a combination of parameters, activations, activation gradients, and parameter gradients. The memory for parameters is equal to the storage of parameters. We estimate the memory for activations by taking the maximum value of multiple measurements. For simplicity, we set the memory for gradients of activations to be equal to the memory for activations. We do not consider the memory for hyper-parameters and momentum. Assuming the memory for dense and sparse parameters are $M^p_d$ and $M^p_s$ respectively, and the memory for dense and sparse activations are $M^a_d$ and $M^p_s$, the training memory for each algorithm would be:
\begin{itemize}
    \item \textbf{FedAVG.} This technique requires the training of a dense model; thus, the memory for the gradients of parameters is close to $M^p_d$. The memory footprint for training is approximately $2M^p_d + 2M^a_d$.
    \item \textbf{FL-PQSU.} This technique trains a static sparse model, so the memory for parameter gradients is close to $M^p_s$. The memory needed for training is approximately $2M^p_s + 2M^a_d$.
    \item \textbf{FedTiny and FedDST.} Since these methods only update the TopK gradients in memory to adjust the model structure, extra memory is used to store the top-$\xi$ gradients and their indices. Therefore, the memory for the parameter gradients is approximate $M^p_s + M_\xi$, where $ M_\xi$ is the memory for the TopK gradients. Consequently, the total memory footprint is $2M^p_s + 2M^a_d + M_\xi$.
    \item \textbf{FedMef.} In comparison to FedTiny and FedDST, FedMef applies scaled activation pruning to activation, resulting in a cache memory of activation of $M^a_d$. However, the activation gradients are not pruned, leading to a total memory footprint of $2M^p_s + M^a_s +M ^a_d +  M_\xi$.
\end{itemize}

\subsection{Training FLOPs}
Compared to other baselines, FedMef incurs minimal computational overhead. Firstly, in budget-aware extrusion, the computational overhead is attributed to the calculation of the regularization term $||\theta_{low}||$, with a complexity of $O(|\theta_{low}|) = O(|\theta|)$. Second, in Scaled Activation Pruning, the computational overhead arises from the normalization in Normalized Sparse Convolution and activation pruning. The normalization operation applies only to unpruned parameters, resulting in a computational complexity of $O((1-s_{m})|\theta|) = O(|\theta|)$. Additionally, the complexity associated with activation pruning is denoted as $O(|a|\log |a|)$, where $|a|$ represents the number of activation elements. The cumulative computational overhead is thus defined as $O(|\theta| + |a| \log |a|)$. These computational overheads are considered negligible compared to the intricate computations during training.

Training FLOPs comprise both forward pass FLOPs and backward pass FLOPs, where the total operations are tallied layer by layer. In the forward pass, layer activations are computed sequentially using previous activations and layer parameters. During the backward pass, each layer computes the activation gradients and the parameter gradients, assuming \textbf{twice} as many FLOPs in the backward pass as in the forward pass. FLOPs in batch normalization and loss calculation are omitted.

In detail, assuming that the inference FLOPs for dense and static sparse models are $F_d$ and $F_s$, and the local iteration number is $E$, the maximum training FLOPs for each framework are as follows:

\begin{itemize}
\item \textbf{FedAVG.} Necessitates training a dense model, resulting in training FLOPs per round equal to $3F_dE$.
% \item \textbf{FL-PQSU.} Trains a static sparse model, with a maximum of training FLOPs equal to $3F_sE$.
\item \textbf{FedTiny and FedDST.} Utilizes RigL-based methods to update model architectures, requiring clients to calculate dense gradients in the last iteration. The maximum training FLOPs are $3F_s(E-1) + F_s + 2F_d$.
\item \textbf{FedMef.} Compared to FedTiny and FedDST, FedMef incurs a slight calculation overhead for BaE and SAP. Therefore, the maximum training FLOPs are $3(F_s+F_o)(E-1) + (F_s+F_o) + 2F_d$, where $F_o$ is the computing overhead of BaE and SAP. We estimate $F_o$ as $F_o = 4(1-s_m)n_\theta + n_a\log n_a$, where $n_\theta$ is the number of parameters $\theta$ and $n_a$ is the number of activation elements. $4(1-s_m)n_\theta$ represents the FLOPs of regularization and WSConv, while $n_a\log n_a$ denotes the FLOPs for activation pruning.
\end{itemize}

\subsection{The Communication Cost}
Regarding the communication costs, FedMef aligns with the communication cost of FedTiny~\cite{huang2022fedtiny}. In contrast to other baselines such as FedDST, wherein, for every $\Delta R$ rounds, clients are required to upload the TopK gradients to the server to support parameter growth, the number of TopK gradients $\xi_t$ is aligned with the count of marked parameters $\theta_{low}$, where $\xi_t = \zeta_t(1-s_m)n_\theta$ and $\zeta_t = 0.2(1 + cos\frac{t\pi}{R_{stop}E})$ is the adjustment rate for the $t$-th iteration. Consequently, the upload overhead is minimal. Furthermore, there is no communication overhead for the model mask $m$ during download because sparse storage formats, such as bitmap and coordinate offset, contain identical element position information. We omit other auxiliary data, such as the learning rate schedule. 

Therefore, assuming that the storage for dense and sparse parameters is $O_d$ and $O_s$, respectively, the data exchange per round is:
\begin{itemize}
\item \textbf{FedAVG.} The data exchange is $2O_d$, containing uploading and downloading dense parameters.
\item \textbf{FedDST.} As mentioned above, the model mask does not require extra space to store, as the compressed sparse parameters already contain the mask information. Therefore, the data exchange per round is $2O_s$, including uploading and downloading sparse parameters.
\item \textbf{FedTiny and FedMef.} Compared to FL-PQSU and FedDST, FedTiny and FedMef require uploading TopK gradients every $\Delta R$ rounds. Therefore, the maximum data exchange per round is $2O_s + O_\xi$, where $O_\xi$ denotes the storage of the TopK gradients.
\end{itemize}

\section{More Experiments}
To showcase the efficiency of the proposed FedMef, we conducted experiments in various federated learning scenarios. Moreover, we also analyze the sensitivity to key hyperparameters in the proposed FedMef. Additionally, to demonstrate the efficacy of FedMef in various model architectures, we selected ResNet34 and ResNet50 for experimentation.

\subsection{Impact of Local Epochs Number}
Due to resource constraints and limited device battery life, the number of local training epochs is necessarily restricted. However, this constraint may impact the training of federated pruning frameworks and potentially undermine their performance. To assess this, we evaluate FedMef and other baseline frameworks on the CIFAR-10 dataset under varying local epoch numbers, employing the ResNet34 model.

As depicted in Table~\ref{tb:E}, our findings reveal that a smaller number of local epochs can affect the performance of FedAVG, and this impact extends to the federated pruning framework as well. Nevertheless, FedMef consistently outperforms other baselines. Notably, the performance of FedDST experiences a significant decline when the local epoch number is very low, underscoring the necessity of sufficient local training for FedDST before adjusting model architecture.

\begin{table}[t]
    \centering
    \begin{tabular}{c|ccc}
    \hline
     $\#$ Epochs  & 10 & 5 & 2 \\ \hline
      FedAVG  & 82.32\% & 79.87\% & 70.15\%  \\
      \hline
     FedDST &  80.30\% & 76.04\% & 64.70\% \\
     FedTiny & 80.41\% & 77.35\% & 70.01\% \\
     \textbf{FedMef} & \textcolor{red}{\textbf{81.18\%}} & \textcolor{red}{\textbf{78.46\%}}& \textcolor{red}{\textbf{70.06\%}}\\
    \hline
    \end{tabular}
    \caption{ Mean accuracy of FedMef and baseline frameworks on CIFAR-10 with various numbers of local epochs per round using ResNet34 model.}
    \label{tb:E}
\end{table}

\begin{table}[t]
    \centering
    \begin{tabular}{c|cccc}
    \hline
     $\#$ Clients  & 10 & 5 & 3 & 1 \\ \hline
      FedAVG  & 78.25\% & 77.47\% & 74.94\%  & 66.38\% \\
      \hline
     FedDST &  72.28\% & 73.21\% & 68.26\%  & 56.05\%\\
     FedTiny & 74.42\% & 74.52\% & 72.36\%  & \textcolor{red}{\textbf{61.45\%}} \\
     \textbf{FedMef} & \textcolor{red}{\textbf{76.24\%}} & \textcolor{red}{\textbf{76.33\%}}& \textcolor{red}{\textbf{72.41\%}}  & 61.04\%\\
    \hline
    \end{tabular}
    \caption{ Mean accuracy of FedMef and baseline frameworks on CIFAR-10 with various numbers of selected clients per round using ResNet50 model. }
    \label{tb:clients}
\end{table}

\subsection{Impact of Selected Clients Number}
Due to diverse network conditions on devices, the number of clients participating in each round is limited. However, this constraint, while improving the negative effect of data heterogeneity, can slow down the convergence speed and affect final performance. In our evaluation, we evaluate the proposed FedMef and other baseline frameworks with various numbers of selected clients per round, utilizing the ResNet50 model.

As presented in Table~\ref{tb:clients}, FedMef consistently outperforms other baselines in most cases. Notably, the performance of FedDST decreases significantly compared to our FedMef and FedTiny, underscoring the necessity of sufficient local training for FedDST.

\subsection{Impact of Regularization Coefficient $\lambda$}

\begin{figure}[t]
  \centering
  \includegraphics[width=0.7\linewidth]{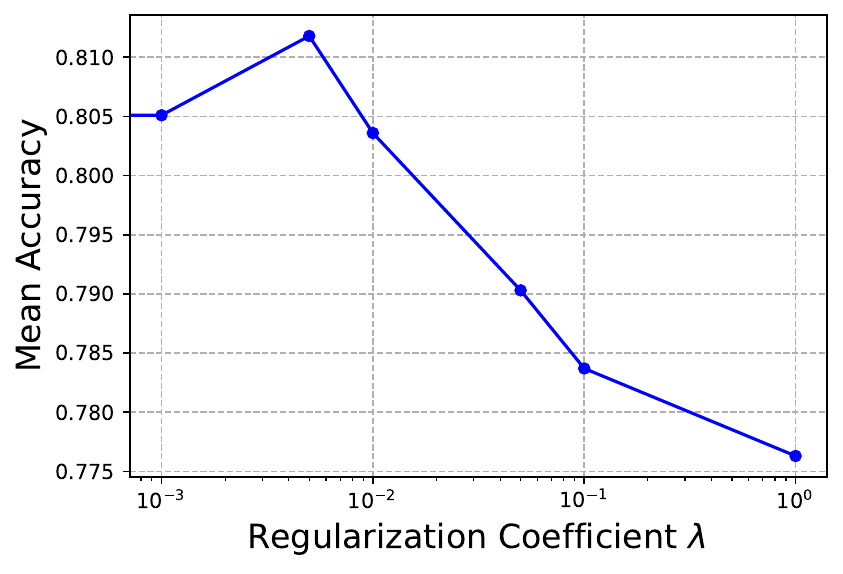}
  \caption{The mean accuracy of the proposed FedMef with various regularization coefficient $\lambda$}
  \label{fig:lam}
\end{figure}

We assess the sensitivity of the regularization coefficient ($\lambda$) in the proposed budget-aware extrusion (BaE). Different coefficients of $\lambda$ are set in FedMef and experiments are conducted on the CIFAR-10 dataset using the ResNet34 model.

As illustrated in Figure~\ref{fig:lam}, the accuracy of FedMef initially increases and then decreases sharply as the coefficient $\lambda$ increases. The initial increase demonstrates the effectiveness of budget-aware extrusion, while the subsequent decrease is attributed to large $\lambda$ values that rapidly zero out the parameters, resulting in an excessively sparse model.

\subsection{Empirical comparison with dense model}
To further demonstrate the effectiveness of BaE and SAP, we conducted experiments on both dense and sparse models on the CIFAR-10 dataset. The results, as illustrated in Figure~\ref{tb:dense}, indicate that SAP maintains the performance of dense models effectively, while reducing the memory footprint of activation. Furthermore, FedMef with BaE outperforms its counterpart without BaE (FedAVG+SAP), underscoring BaE's contribution to improving performance.

\begin{table}[t]
\resizebox{\columnwidth}{5mm}{
\begin{tabular}{c|lll}
\hline
\multicolumn{1}{l|}{Acc(Memory)} & FedMef($s_{tm}=0.8$) & FedAVG + SAP & FedAVG \\ \hline
MobileNetV2 & 65.50\%(84.94 MB) & 64.04\%(101.28 MB) & 64.28\%(148.63 MB) \\ \hline
ResNet18 & 81.73\%(45.91 MB) & 81.32\%(111.51 MB) & 81.15\%(120.74MB) \\ \hline
\end{tabular}
}
\vspace{-0.2cm}
\caption{\hong{Accuracy and memory footprint on the CIFAR-10 dataset}}
\label{tb:dense}
\vspace{-0.5cm}
\end{table}

\section{More Detail and Information}
\subsection{Hardware information}
\hong{We follow most FL pruning work to simulate training on GPUs, while now deploying on a Raspberry Pi 4B with 1GB RAM}

\subsection{NSConv v.s. BN} 
\hong{NSConv is more suitable for CNNs, which are popular on edge devices. It excels with sparse weights, as it may introduce more computational overhead on dense weights. NSConv outperforms BN when the batch size is small, as shown in Figure~\ref{fig:other_1} right(FedMef v.s. FedMef w/o SAP). This is important for low-memory devices that can only train with a small batch size. NSConv matches BN performance when batch size is large, as shown in Figure~\ref{fig:hist} .}

\subsection{Structured pruning v.s. Unstructured pruning}
\hong{We use unstructured pruning(pruning parameters) instead of structured pruning (pruning filters) as structured pruning often suffers form a serious performance drop when the sparsity is higher than $10\%$~\cite{shen2022prune} and has a limited impact on memory reduction. In contrast, our proposed unstructure pruning method can achieve 80\%+ sparsity while maintaining accuracy, as shown in Table~\ref{tb:cost}.}

% \section{Less information loss in Budget-aware Training}
% \label{sec:less}

% \begin{figure*}[h]
%   \centering
%   \includegraphics[width=0.45\linewidth]{Image/info_loss.pdf}
%   \hspace{0in}
%   \includegraphics[width=0.45\linewidth]{Image/converge.pdf}
%   \caption{\textit{Left}: the mean magnitude of pruned parameters in FedMef and the SOTA baseline, FedTiny~\cite{huang2022fedtiny}, framework. \textit{Right}: the validation accuracy during the training in FedMef and baseline framework. FedMef framework is much more stable than baseline method.}
%   \label{fig:motivation}
% \end{figure*}
% \hspace{-1cm}

\end{document}